\documentclass[10pt,twocolumn]{article}
\usepackage[letterpaper,textwidth=6.875in,textheight=8.875in,top=1in,columnsep=0.3125in]{geometry}

\usepackage[T1]{fontenc}
\usepackage[utf8]{inputenc}
\usepackage{mathptmx}
\usepackage{amsmath,amssymb,booktabs,graphicx,array}
\graphicspath{{figures/}}
\usepackage{xcolor}
\usepackage[hidelinks]{hyperref}
\makeatletter
\renewcommand{\@seccntformat}[1]{\csname the#1\endcsname.\hspace{0.5em}}
\renewcommand{\section}{\@startsection{section}{1}{\z@}
  {-2.0ex plus -0.5ex minus -0.2ex}{0.9ex plus 0.2ex}
  {\normalfont\fontsize{12}{14}\selectfont\bfseries}}
\renewcommand{\subsection}{\@startsection{subsection}{2}{\z@}
  {-1.7ex plus -0.5ex minus -0.2ex}{0.7ex plus 0.2ex}
  {\normalfont\fontsize{11}{13}\selectfont\bfseries}}
\makeatother
\renewenvironment{abstract}{%
  \begin{center}\fontsize{12}{14}\selectfont\bfseries Abstract\end{center}%
  \itshape\noindent\ignorespaces
}{\par\normalfont\vspace{0.8em}}

\newcommand{\papertitle}{EmbodyAgent: Post-Training Embodied VLM Agents via Distillation, SFT, and RL}
\newcommand{\RLSuccesses}{84}
\newcommand{\RLBestRound}{64}
\newcommand{\RLSamplingStart}{8.4}
\newcommand{\RLSamplingEnd}{87.7}
\newcommand{\paperauthors}{Wei Dai\textsuperscript{1*}\quad Jinkun Liu\textsuperscript{1*}\quad Jiayi Chen\textsuperscript{2}\quad Han Liu\textsuperscript{1}\quad Wenbo Ding\textsuperscript{1\ensuremath{\dagger}}\quad Shaolun Huang\textsuperscript{1\ensuremath{\dagger}}}
\newcommand{\paperaffiliation}{$^{\mathrm{1}}$Tsinghua University\quad$^{\mathrm{2}}$Xidian University}

\newcommand{\paperemails}{\{daiw26, liujk22, liuhan26\}@mails.tsinghua.edu.cn\quad chenjy26@stu.xidian.edu.cn}
\newcommand{\papercorrespondingemails}{\{ding.wenbo, shaolun.huang\}@sz.tsinghua.edu.cn}

\hypersetup{pdftitle={\papertitle},pdfauthor={Wei Dai, Jinkun Liu, Jiayi Chen, Han Liu, Wenbo Ding, and Shaolun Huang}}
\begin{document}
\twocolumn[\vspace*{0.5em}
\begin{center}
\fontfamily{ptm}\selectfont
{\fontsize{14}{17}\selectfont\bfseries \papertitle\par}
\vspace{1.6em}
{\fontsize{12}{14}\selectfont \paperauthors\par}
\vspace{0.5em}
{\fontsize{12}{14}\selectfont \paperaffiliation\par}
\vspace{0.4em}
{\small\texttt{\paperemails}\par}
{\small\texttt{\papercorrespondingemails}\par}
\end{center}
\vspace{0.6em}
]
\begingroup
\renewcommand{\thefootnote}{\fnsymbol{footnote}}
\footnotetext[1]{Equal contribution.}
\footnotetext[2]{Co-corresponding authors.}
\endgroup
\begin{abstract}
Vision-language models (VLMs) can act as closed-loop \emph{robot-use agents} (RUAs) by writing and executing control code through tool interfaces\cite{isola2026robotuse,yuan2026openrua}. However, turning an open-weight VLM into a high-performing embodied agent requires a complete post-training stack: out-of-the-box open VLMs fail completely at interactive tool contracts, strong frontier teachers are typically accessible only as black-box endpoints without exposed weights or token probabilities, and supervised imitation alone plateaus under rollout distribution shift. We present \textbf{EmbodyAgent}, an end-to-end post-training framework for embodied VLM agents built around three tightly coupled pipelines: (1) a \emph{black-box data synthesis pipeline} that queries a strong teacher model---requiring no teacher weights or internal signals---under a forced \emph{think-then-code} contract and filters trajectories by task success predicates; (2) an \emph{SFT pipeline} that distills these explanation-and-program targets into an open-weight 9B VLM with both vision and language encoders kept trainable; and (3) an \emph{online RL pipeline} that optimizes student-generated episodes via group-relative advantage using a runtime-error-penalized outcome reward. All three pipelines share a single primitive-free Python execution tool exposing only raw visual observation and seven-dimensional end-effector control, requiring no action decoder and no interface shift across stages. Under a strictly separated evaluation protocol on ten LIBERO Spatial tasks\cite{liu2023libero}---disjoint initial-state splits, greedy decoding at test time, and training-only safeguards---held-out success on 100 test states rises from 0/100 (base 9B VLM) to 14/100 (after SFT) and \RLSuccesses/100 (after RL), while RL training rollout success climbs from \RLSamplingStart\% to \RLSamplingEnd\% and median model calls per test episode drop from 36 to 11. We further identify a self-reinforcing runtime-error mode induced by outcome-based RL on executable code, and establish the failure-category monitoring and sampling-side curation protocol that stabilizes training.
\end{abstract}
\section{Introduction}
Closed-loop robot manipulation demands tight coordination between perception, planning, and control: an agent must decide how to structure control commands over time, when acquiring a fresh observation warrants another model turn, and how to recover when execution diverges from expectation. Recent \emph{robot-use agent} (RUA) systems~\cite{isola2026robotuse,yuan2026openrua,chu2026openrua} demonstrate that vision-language models (VLMs) can supply this coordination directly through tool use~\cite{zhang2026unexpected,si2026fewertokens,qu2026roboharness,li2026k1,chen2026showharness}, extending the interaction paradigm of digital computer-use agents~\cite{agashe2025agents2,wang2025uitars2} to physical environments. Yet equipping an open-weight VLM with reliable closed-loop manipulation capability is not a prompting problem---it is a full-pipeline post-training problem. Off-the-shelf open VLMs fail completely under interactive control contracts (0/100 held-out tasks for an untrained 9B model); frontier models capable of solving these tasks zero-shot are often accessible only as black-box endpoints without exposed weights, hidden states, or token logits; and supervised fine-tuning (SFT) on teacher trajectories alone stalls early (14/100) because an imitated policy encounters its own execution histories, errors, and observations at test time rather than the teacher's.

To close this end-to-end gap, we build \textbf{EmbodyAgent}, a complete post-training stack for embodied VLM agents organized as three successive pipelines:
\begin{enumerate}
\item \textbf{Black-box strong-teacher data synthesis.} A frontier VLM teacher interacts with the environment strictly as a black-box agent through the tool harness---requiring zero access to teacher weights or logits. Our synthesis pipeline enforces structured reasoning before every code output and filters episodes by environment success predicates, turning raw black-box rollouts into clean multimodal training trajectories.
\item \textbf{Multimodal SFT distillation.} An open-weight 9B VLM student is distilled on the synthesized explanation-and-program targets with both its vision encoder and language backbone kept trainable. This stage eliminates cold-start format failures and grounds visual observations into executable control loops and observation requests.
\item \textbf{Online outcome-based RL.} Starting from the SFT checkpoint with no teacher in the loop, group-relative reinforcement learning optimizes the student VLM directly on its own sampled episodes using a runtime-error-penalized success reward. Online optimization overcomes imitation distribution shift and substantially improves both task success and multi-turn decision efficiency.
\end{enumerate}

All three pipelines are unified by a single \emph{think-then-code} policy representation over one Python execution tool (\texttt{execute\_python}). Unlike prior code-as-policy agents~\cite{liang2022code,ahn2022saycan} that compose a hand-engineered library of manipulation primitives, EmbodyAgent exposes only two raw runtime operations: \texttt{env.observation()}, which returns primary and wrist camera images together with robot proprioception, and \texttt{env.action(a)}, which advances one simulator control step with $a\in[-1,1]^7$. At every turn, the VLM is required to emit a chain-of-thought explanation planning loop structure, step batching, and observation timing, followed by an executable Python program. Because the policy lives entirely within a standard multimodal text-and-code interface---with no external action decoder and no pre-defined primitives---black-box synthesis, SFT, and RL operate over the exact same output contract without any cross-stage representation shift.

We validate the complete pipeline on ten LIBERO Spatial tasks~\cite{liu2023libero} under a strictly separated evaluation protocol: per-task initial states are partitioned into disjoint train, validation, and test splits; Stage~1 synthesizes 100 successful episodes (1{,}475 targets) from the black-box teacher Astra Medium on train states; Stage~2 distills Qwen3.5-9B~\cite{yang2025qwen3} across vision and language parameters; and Stage~3 runs temperature-one group-relative RL on train and validation states while evaluating greedily on all 100 held-out test states after every round under full episode budgets and zero test-time safeguards. Across this progression, held-out success advances from 0/100 (base 9B VLM) to 14/100 (after SFT) and \RLSuccesses/100 (after \RLBestRound{} rounds of RL), while RL training rollout success rises from \RLSamplingStart\% to \RLSamplingEnd\% with all ten tasks improving. Simultaneously, Stage~3 RL sharpens the VLM's execution efficiency, compressing median model calls per test episode from 36 (SFT) to 11 and yielding a self-contained 9B VLM that executes locally with no ongoing reliance on the black-box teacher.

Executing outcome-based RL over multi-turn code generation also surfaces a failure mode absent from fixed-action policies: a self-reinforcing runtime-error loop in which execution tracebacks enter the multimodal history that subsequent turns condition on, allowing error-prone patterns to amplify before recovery. We show that aggregate success rate lags this pathology and establish a per-round failure-category monitoring and sampling-only curation protocol that restores healthy optimization without altering evaluation.

Our contributions are threefold:
\begin{itemize}
\item \textbf{A full-pipeline Embodied Agent post-training stack.} An end-to-end framework integrating weight-free black-box teacher data synthesis, multimodal SFT distillation, and online outcome-based RL that trains an open-weight 9B VLM from 0/100 to 14/100 to \RLSuccesses/100 across ten LIBERO Spatial tasks.
\item \textbf{A unified think-then-code interface across stages.} A single primitive-free execution contract (\texttt{env.observation()}, \texttt{env.action(a)}) shared verbatim by synthesis, SFT, and RL, turning action composition and observation scheduling into explicit VLM learning targets with no action decoder or cross-stage mismatch.
\item \textbf{Diagnosis and stabilization of executable-code RL.} Identification of a self-reinforcing runtime-error loop induced by outcome optimization over execution feedback, paired with failure-category monitoring and training-side safeguards that stabilize RL without contaminating test evaluation.
\end{itemize}

\section{Related Work}

\paragraph{Robot-use agents and code-as-policy.}
Code as Policies\cite{liang2022code} established that language models can close the control loop on robots by emitting programs over perception and actuation APIs, followed by SayCan\cite{ahn2022saycan}, Inner Monologue\cite{huang2022inner}, ProgPrompt\cite{singh2023progprompt}, and VoxPoser\cite{huang2023voxposer} grounding code in affordances, feedback, structured prompts, and 3D value maps. Executable code also strengthens general language agents\cite{wang2024codeact}, and autonomous coding-agent loops\cite{jimenez2024swebench,yang2024sweagent} have extended into physical manipulation\cite{li2026embodiedevo} under the umbrella of the \emph{robot-use agent} (RUA)---a VLM operating a robot through a tool and feedback harness\cite{isola2026robotuse,yuan2026openrua,chu2026openrua}. Recent harness works equip frozen frontier VLMs with perception, motion, and geometric tooling\cite{qu2026roboharness,li2026k1,chen2026showharness,guo2026robodawn,zhang2026unexpected,si2026fewertokens}, where generated code selects which pre-built routine to invoke. Concurrent post-training efforts explore complementary directions: CaPtain\cite{fan2026captain} post-trains an open model over guarded motion primitives using per-request credit from a privileged simulator critic, whereas EmbodiedEvo\cite{li2026embodiedevo} evolves external skills and lessons in context without updating model weights. In contrast, EmbodyAgent targets complete, weight-updating post-training of an open-weight VLM over raw observations and seven-dimensional control alone: it requires no motion-primitive library, no teacher weights at data synthesis, and no privileged simulator critic during RL.

\paragraph{Reasoning and distillation for interactive agents.}
Chain-of-thought prompting\cite{wei2022cot}, outcome-filtered bootstrapping\cite{zelikman2022star}, and outcome-based reinforcement\cite{deepseek2025r1} demonstrate the value of explicit reasoning traces, while interleaved reasoning and tool calling\cite{yao2023react,schick2023toolformer} and reasoning-trace distillation\cite{hsieh2023distilling} transfer interactive problem-solving from strong teachers into smaller open models. In embodied learning, Embodied Chain-of-Thought\cite{zawalski2024ecot} and CoT-VLA\cite{zhao2025cotvla} pair reasoning with action-decoder policies ahead of single actions or short chunks. Under EmbodyAgent's think-then-code contract, the VLM's explanation plans an entire multi-step program and its observation requests, allowing reasoning, control composition, and observation scheduling to be distilled together in SFT and refined together in RL.

\paragraph{VLM-based embodied policies and VLAs.}
A large body of work adapts VLMs to manipulation via continuous or tokenized action decoders, including PaLM-E\cite{driess2023palme}, RT-1 and RT-2\cite{brohan2022rt1,brohan2023rt2}, OpenVLA\cite{kim2024openvla}, Octo\cite{octo2024}, $\pi_0$\cite{black2024pi0}, RDT-1B\cite{liu2024rdt}, and RoboFlamingo\cite{li2023roboflamingo} on LIBERO, with SimpleVLA-RL\cite{li2025simplevla} adding RL and Harness VLA\cite{zhang2026harnessvla} wrapping a frozen VLA inside an agent harness. Rather than attaching a separate action head, EmbodyAgent post-trains a VLM directly inside its native multimodal text-and-code interface, where one model turn can express loops, conditionals, and explicit observation requests. The two paradigms are orthogonal---a post-trained code-writing VLM agent could invoke an action-decoder VLA as a subroutine---and our focus here is establishing the full-pipeline post-training recipe (black-box synthesis, SFT, and RL) for executable VLM policies.

\paragraph{Outcome-based RL for executable policies.}
PPO\cite{schulman2017ppo}, value-free group-relative optimization in GRPO\cite{shao2024grpo}, and asymmetric clipping in DAPO\cite{yu2025dapo} provide the algorithmic basis for outcome-driven post-training of reasoning models\cite{deepseek2025r1} without process supervision\cite{lightman2024verify}, while execution feedback has been used to optimize code generation\cite{le2022coderl} and agent reflection\cite{shinn2023reflexion}. Applying outcome-based RL to multi-turn embodied code generation exposes a distinct instability: runtime-error feedback enters the context that subsequent turns condition on, creating a self-reinforcing error basin that we diagnose via failure-category monitoring and resolve with training-side curation.

\section{The EmbodyAgent Post-Training Pipeline}
\label{sec:method}
\begin{figure*}[t]
\centering
\includegraphics[width=0.99\textwidth]{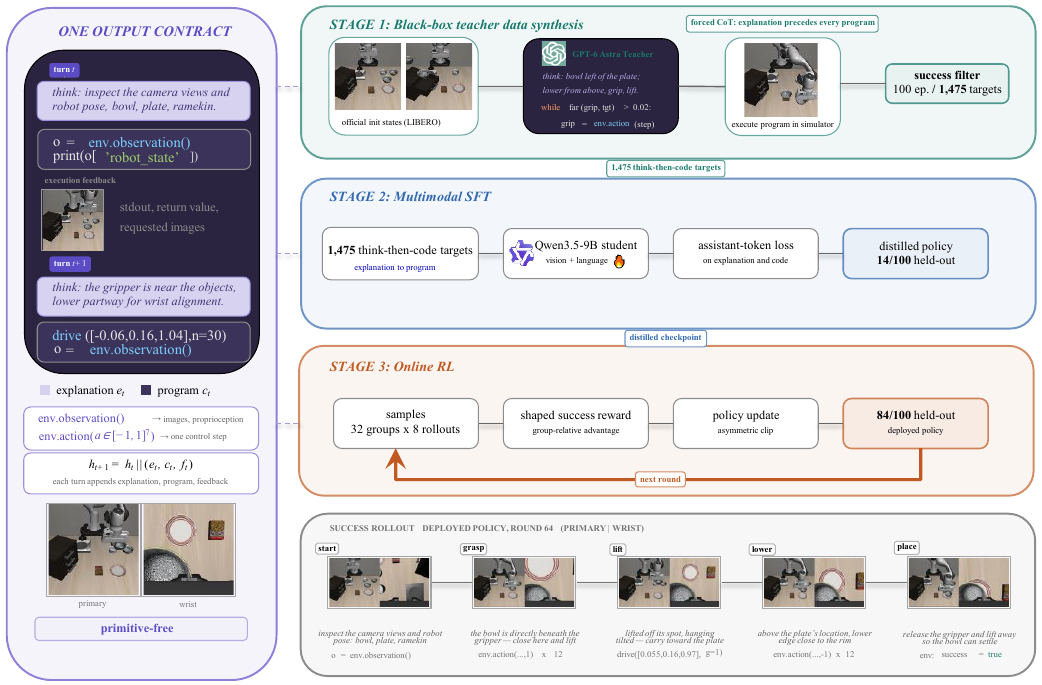}
\caption{The EmbodyAgent full-pipeline post-training framework; the photographs are LIBERO Spatial frames. \emph{Left:} the single output contract shared verbatim across all stages---at every turn the VLM emits a forced explanation $e_t$ followed by an executable program $c_t$ over two raw runtime operations, shown for two turns of the episode below. \emph{Center:} the three-stage pipeline. Stage~1 (black-box data synthesis) queries Astra Medium on training-split initial states without accessing teacher weights and filters successful think-then-code episodes into 1{,}475 targets; Stage~2 (multimodal SFT) distills these targets into an open-weight 9B VLM; Stage~3 (online RL) optimizes the student's own rollouts via group-relative outcome credit. \emph{Bottom:} a successful held-out episode from the post-trained 9B VLM (RL round \RLBestRound), pairing each frame with its verbatim explanation and program through grasp, lift, lower, and place.}
\label{fig:pipeline}
\end{figure*}

Figure~\ref{fig:pipeline} illustrates the complete EmbodyAgent post-training stack: a unified executable interface shared verbatim across three successive pipelines---Stage~1 black-box teacher data synthesis, Stage~2 multimodal SFT, and Stage~3 online RL (detailed in Section~\ref{sec:rl}). Each pipeline consumes the preceding stage's outputs without altering the model's observation, reasoning, or action schema.

\subsection{Shared think-then-code policy interface}
We formulate closed-loop manipulation as a multi-turn partially observed interaction conditioned on a language instruction $g$. At model turn $t$, the multimodal context $h_t$ contains $g$, previous assistant turns, and environment execution feedback, including any explicitly requested images. Rather than emitting raw action vectors, the VLM policy outputs a public chain-of-thought explanation $e_t$ followed by an executable Python program $c_t$:
\begin{equation*}
 \pi_\theta(e_t,c_t\mid h_t)
 =\pi_\theta(e_t\mid h_t)\pi_\theta(c_t\mid h_t,e_t).
\end{equation*}

A persistent Python runtime executes $c_t$ inside a minimal environment interface exposing only two operations: \texttt{env.observation()}, which returns primary and wrist camera images together with robot proprioception, and \texttt{env.action(a)}, which steps the existing simulator controller with a seven-dimensional end-effector command $a\in[-1,1]^7$ (translation, rotation, and gripper). Unlike code-as-policy frameworks that select from a pre-built library of manipulation primitives, this interface provides no primitives: the VLM synthesizes its own control loops, conditionals, and action batches directly over raw sensing and actuation.

Executing $c_t$ returns feedback $f_t$---stdout, exceptions, the last expression's value, and any requested images---and advances the history:
\begin{equation*}
 h_{t+1}=h_t\mathbin{\Vert}(e_t,c_t,f_t).
\end{equation*}
Because visual observations enter the context only when $c_t$ invokes \texttt{env.observation()}, a single program can execute multiple control steps before consulting the VLM again. This keeps the entire policy inside a standard multimodal text-and-code interface with no external action decoder, making both action composition and observation scheduling directly trainable across synthesis, SFT, and RL.

\subsection{Stage 1: Black-box teacher data synthesis}
The goal of Stage~1 is to synthesize high-quality multimodal training trajectories from a strong teacher VLM without requiring access to teacher weights, activations, or token log-probabilities. We deploy Astra Medium (a frontier VLM at medium reasoning effort) strictly as a black-box agent operating through the \texttt{execute\_python} harness on official pre-generated LIBERO initial states restricted to the training split.

Left unconstrained, the teacher defaults to emitting bare code blocks without explicit planning. The synthesis harness therefore enforces the think-then-code contract at every turn: any response omitting the explanation terminates the episode immediately as a format failure and is discarded. Completed episodes are retained if and only if the environment's BDDL success predicate passes. Across 1{,}620 synthesis attempts (32\% task-averaged acceptance rate over valid attempts), Stage~1 collects 100 validated episodes---ten per task across the ten LIBERO Spatial tasks---and exports them into 1{,}475 multimodal assistant-response targets. Each target pairs the full interaction history with the teacher's explanation and executable program, providing clean supervision for control synthesis, formatting, and observation timing purely from black-box interaction.

\subsection{Stage 2: Multimodal supervised fine-tuning}
An out-of-the-box open-weight 9B VLM fails completely on the interactive contract (0/100 held-out states), making supervised cold-start distillation essential before online exploration. Let $Y_i$ denote the explanation, code, and output-format tokens in synthesized target $i$. Stage~2 optimizes the standard autoregressive cross-entropy objective over assistant outputs:
\begin{equation*}
 \mathcal L_{\mathrm{SFT}}=-\frac{1}{\sum_i|Y_i|}
 \sum_{i}\sum_{j\in Y_i}\log\pi_\theta(y_{ij}\mid h_i,y_{i,<j}),
\end{equation*}
while masking prompt history, visual inputs, and tool feedback from the loss. Supervision preserves the exact explanation-then-program generation order enforced during Stage~1. Crucially, both the vision encoder and the language parameters of the 9B VLM are kept trainable so visual grounding and program synthesis co-adapt. After ten epochs of SFT, we verify on held-out states that the checkpoint produces format-valid outputs and well-formed tool calls, and advance it directly into Stage~3 RL under the unchanged contract.

\section{Stage 3: Outcome-Based Reinforcement Learning}
\label{sec:rl}
Stage~1 black-box synthesis and Stage~2 SFT equip the 9B VLM with format compliance and initial closed-loop manipulation skill (0/100 $\to$ 14/100). However, SFT supervises only trajectories generated by the teacher, whereas at rollout time the student VLM must condition on its own generated programs, execution feedback, and resulting visual observations. Stage~3 closes this gap through online outcome-based reinforcement learning on student-generated episodes, requiring no teacher supervision, no learned critic, and no change to the think-then-code interface.

\subsection{Group-relative outcome optimization}
Training proceeds in rounds. In each round, the current VLM policy samples groups of $G=8$ episodes on initial states drawn exclusively from the training and validation splits of our state manifest---never from held-out test states. Each episode receives a terminal outcome reward built around the environment's BDDL success predicate, where $k_i$ counts the cells in episode $i$ whose execution raised a runtime error:
\begin{equation*}
 R_i=\begin{cases}
  1.0-0.05\,k_i, & \text{success (BDDL predicate passes)}\\[2pt]
  -0.05\,k_i, & \text{other failure (floor $-1.0$)}\\[2pt]
  -0.2, & \text{failure ending within 50 simulator steps.}
 \end{cases}
\end{equation*}
This shaping preserves strict outcome ordering: every success scores at least $0.8$ and thus outscores every failure ($\le 0$), error-heavy failures rank below clean attempts, and the fast-fail floor prevents the policy from learning to quit early. The reward function remains fixed throughout training. Within each group, episode $i$ receives the group-relative advantage
\begin{equation*}
 A_i=\frac{R_i-\overline R}{s(R)+\varepsilon},
\end{equation*}
where $\overline R$ and $s(R)$ are the group mean and sample standard deviation and $\varepsilon$ is a small constant. Groups whose $G=8$ episodes share identical binary outcomes carry no success contrast and are discarded; sampling continues until 32 informative groups are gathered (capped at 160 attempted groups per round). This filtering acts as an implicit curriculum, concentrating gradient updates on initial states where the current policy is stochastic. Episodes failing due to infrastructure errors unrelated to the policy are resampled (up to three times) and excluded from reward statistics. Following GRPO\cite{shao2024grpo}, our trainer adapts policy-loss primitives from SimpleVLA-RL\cite{li2025simplevla} to variable-length multimodal histories and multi-turn Python responses.

The episode advantage $A_i$ is broadcast across all newly generated explanation and program tokens of episode $i$ (with preceding history and tool feedback masked). Using token importance ratio $\rho_{ij}=\pi_\theta(y_{ij}\mid h_{ij})/\pi_{\mathrm{old}}(y_{ij}\mid h_{ij})$, we optimize the DAPO-style\cite{yu2025dapo} asymmetric clipped surrogate objective:
\begin{equation*}
 \mathcal L_{\mathrm{PG}}=-\mathbb E_{i,j}\!\left[
 \min\!\left(\rho_{ij}A_i,
 \operatorname{clip}(\rho_{ij},1-\varepsilon_{\mathrm{lo}},1+\varepsilon_{\mathrm{hi}})A_i\right)\right],
\end{equation*}
with $(\varepsilon_{\mathrm{lo}},\varepsilon_{\mathrm{hi}})=(0.2,0.28)$. Rollouts are sampled at temperature one from the full vocabulary, storing exact sampled tokens and old-policy log-probabilities for update consistency. Optimization uses AdamW\cite{loshchilov2019adamw} across all VLM parameters---including the vision encoder---under ZeRO-3 sharding\cite{rajbhandari2020zero} on sixteen PPU accelerators with a constant learning rate of $2\times10^{-6}$ and no KL penalty (format violations terminate rollouts in failure and are disciplined directly by the reward). Rollout and evaluation inference is served via vLLM\cite{kwon2023vllm}, and flash-linear-attention kernels accelerate the student's gated-delta-rule layers\cite{yang2024deltanet} during updates.

\subsection{Sampling-side safeguards for executable RL}
Online sampling with an executable Python runtime exposes rollout pathologies absent in static teacher demonstrations. We enforce three safeguards strictly during training sampling:
\emph{(i) Unparseable cells are resampled.} At temperature one, a cell that fails to parse or compile is resampled up to twice before consuming simulator steps; runtime errors are deliberately retained because their tracebacks are genuine policy feedback.
\emph{(ii) Error loops are truncated.} Once an episode accumulates four cells ending in runtime errors, it is terminated early so prolonged traceback--retry loops do not exhaust group sampling budget while the reward's per-error penalty already scores them below honest attempts.
\emph{(iii) Pathological rollouts are replaced.} Failed rollouts whose accumulated context exceeds one million tokens dominate host memory while contributing few informative tokens; they are rejected and resampled (up to three replacements per group), with group advantages recomputed over the final composition.

All three safeguards were active from round~1 and apply \emph{only} during training rollouts. Test evaluation decodes greedily, resamples nothing, truncates nothing, and runs every episode to its full budget or voluntary termination.

\section{Experiments}
\label{sec:experiments}

\subsection{Experimental setup and strict evaluation separation}
All experiments evaluate the ten LIBERO Spatial tasks\cite{liu2023libero} initialized from official pre-generated LIBERO states under a fixed SHA-256 manifest. Per task, states are partitioned into disjoint sets of 30 training, 10 validation, and 10 test states. Stage~1 black-box teacher synthesis samples only from the training split, and Stage~3 RL samples only from the training and validation splits; the 100 held-out test states (ten per task) never enter any training or synthesis stage and are evaluated after every RL round.

Training and evaluation regimes are strictly separated. Stage~3 RL samples at temperature one with the sampling-side safeguards of Section~\ref{sec:rl}; every test evaluation uses greedy decoding, disables all resampling, truncation, and replacement safeguards, and grants the full episode budget (2{,}000 simulator steps, 150 model calls, 600-second code-block timeout) or voluntary termination. Task success is judged solely by the environment's BDDL predicate, and per-episode failure categories are logged on every test evaluation.

\subsection{Stage-by-stage post-training progression}
\begin{figure*}[t]
\centering
\includegraphics[width=0.98\linewidth]{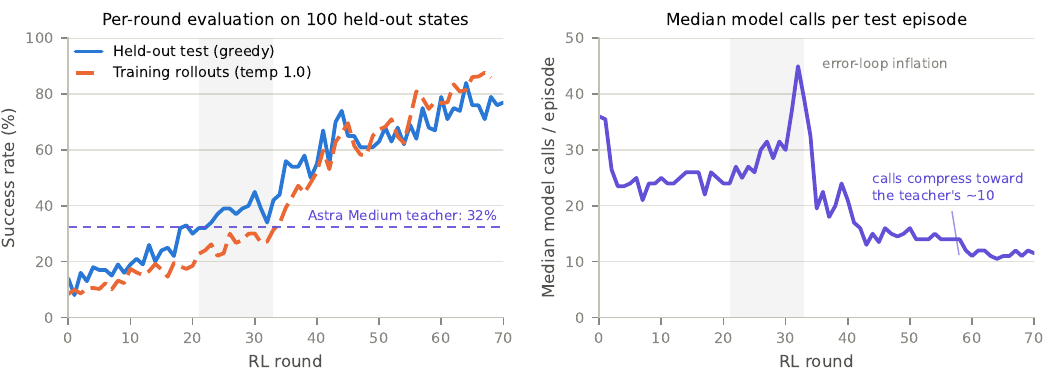}
\caption{Stage~3 RL progression across ten LIBERO Spatial tasks. \emph{Left:} greedy success on the 100 held-out test states after each round alongside temperature-one training rollout success (round~0 is the Stage~2 SFT checkpoint). The purple dashed line indicates the black-box Astra Medium teacher's collection acceptance rate (32\%, validated episodes over valid attempts, averaged over tasks); under RL, student rollout success rises from \RLSamplingStart\% to \RLSamplingEnd\%, surpassing the teacher's collection rate. The shaded band marks rounds 21--33, where runtime-error episodes dominated failures. \emph{Right:} median model calls per test episode, inflating briefly during the error wave (rounds 31--33) and compressing to 10--14 calls once the wave subsides.}
\label{fig:curve}
\end{figure*}

\begin{table}[t]
\centering\small
\caption{Per-task success on the held-out states (ten per task), greedy decoding, identical protocol for both checkpoints (RL at round \RLBestRound).}
\label{tab:comparison}
\begin{tabular}{@{}lrr@{}}
\toprule
Task & SFT & RL \\
\midrule
0 & 2/10 & 10/10 \\
1 & 2/10 & 9/10 \\
2 & 2/10 & 9/10 \\
3 & 2/10 & 9/10 \\
4 & 0/10 & 4/10 \\
5 & 0/10 & 9/10 \\
6 & 0/10 & 9/10 \\
7 & 1/10 & 8/10 \\
8 & 4/10 & 8/10 \\
9 & 1/10 & 9/10 \\
\midrule
Total & 14/100 & 84/100 \\
\bottomrule
\end{tabular}
\end{table}

\begin{figure}[t]
\centering
\includegraphics[width=0.92\linewidth]{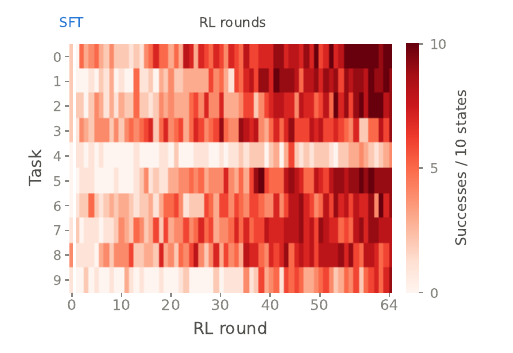}
\caption{Per-task held-out success across Stage~3 RL rounds (column~0 is the Stage~2 SFT checkpoint). Every task improves over SFT, with nine of ten tasks reaching 8--10/10; task~4 remains the hardest throughout, matching the task that also resisted Stage~1 black-box synthesis longest.}
\label{fig:heatmap}
\end{figure}

\begin{figure}[t]
\centering
\includegraphics[width=0.99\linewidth]{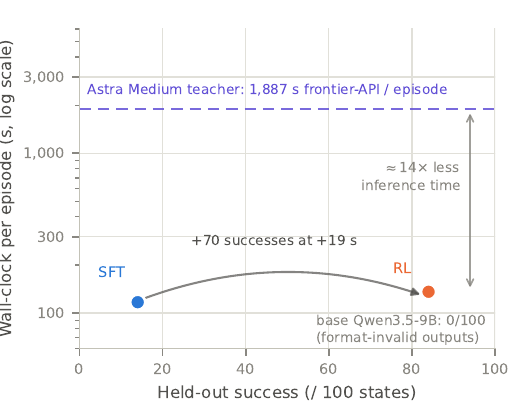}
\caption{Stage-by-stage progression on Qwen3.5-9B (success versus median per-episode wall-clock time). Out of the box, the base 9B VLM solves 0/100 held-out states due to format failure. Stage~2 SFT lifts the same VLM to 14/100 (117\,s/episode), and Stage~3 RL advances it to \RLSuccesses/100 (136\,s/episode, simulation included). The dashed line marks the black-box teacher's 1{,}887\,s of frontier-API inference per episode, paid once offline during Stage~1 synthesis across 1{,}620 attempts.}
\label{fig:cost}
\end{figure}

Table~\ref{tab:comparison} and Figures~\ref{fig:curve}--\ref{fig:cost} quantify what each stage of the post-training pipeline contributes on the identical Qwen3.5-9B\cite{yang2025qwen3} backbone:
\begin{itemize}
\item \textbf{Base 9B VLM (0/100):} Given the exact same tool schema and prompt without post-training, Qwen3.5-9B fails on all 100 held-out states because every episode terminates in a format-invalid output.
\item \textbf{After Stage~2 SFT (14/100):} Distilling the 1{,}475 targets synthesized by the black-box teacher resolves format failure (only five of 86 failures end in a Python parse or runtime error; 85 exhaust the simulator-step budget) and establishes initial closed-loop manipulation.
\item \textbf{After Stage~3 RL (\RLSuccesses/100):} Within \RLBestRound{} rounds of online outcome-based RL, held-out test success climbs by +70 points to \RLSuccesses/100, while temperature-one training rollout success rises from \RLSamplingStart\% to \RLSamplingEnd\%. The close tracking between training rollout success and greedy held-out success confirms genuine generalization beyond the training and validation initial states. All ten tasks improve (Table~\ref{tab:comparison}, Figure~\ref{fig:heatmap}): nine reach 8--10/10, while task~4 rises from 0/10 to 4/10.
\end{itemize}

\subsection{Learned observation scheduling and execution efficiency}
The shared think-then-code representation makes the VLM's observation scheduling directly measurable across stages. In Stage~1 synthesized trajectories, each teacher turn pairs a 38-word explanation with a program of median three lines (mean 3.8, maximum 17), 99.5\% of programs call \texttt{env.observation()}, and an episode averages 14.8 model turns (6--25). After Stage~2 SFT, the student VLM is substantially more verbose in interaction frequency, emitting a median of 36 model calls per test episode (mean 35.2). Stage~3 RL compresses this back without any explicit turn-count penalty: by round~\RLBestRound, the post-trained 9B VLM emits a median of 11 calls per test episode under identical budgets, batching longer control loops per turn while achieving \RLSuccesses/100 success (and 13 of its 16 remaining failures exhaust the 150-call budget rather than simulator steps).

Consequently, Stage~3 RL improves capability and per-episode execution time together (Figure~\ref{fig:cost}): the post-trained 9B VLM completes a successful test episode in a median of 136\,s wall-clock including simulation (versus 117\,s for SFT). Meanwhile, the black-box teacher's 1{,}887\,s of frontier-API inference per episode (72.8\,s per call at the median across 1{,}620 attempts) is incurred solely as a one-time Stage~1 offline synthesis cost, leaving a self-contained open-weight VLM that operates locally with no dependency on teacher weights or API endpoints.

\subsection{Self-reinforcing runtime-error dynamics in Stage 3}
Failure-category monitoring during Stage~3 revealed an optimization pathology specific to outcome-based RL on executable outputs. Through rounds 11--20, held-out runtime-error episodes stay below five per 100. Beginning at round~21, runtime errors surge and plateau at 63--69 per 100 across rounds 24--32, pulling held-out success down from 40 to 34 even as gradient updates continue. Inspection of episode transcripts reveals the feedback loop: a failing cell leaves its error traceback in history $h_t$, subsequent turns condition on that traceback, and trajectories that still pass the success check despite repeated error--retry cycles receive positive group-relative advantage---reinforcing error-generating contexts and code patterns.

Crucially, aggregate success masked this dynamics during its onset, continuing to climb through rounds 21--26 even as runtime errors saturated the failure mix; only per-round failure-category logging exposed it. Unlike an earlier transient spike at rounds 6--7 that dissipated immediately, the round-21 wave persisted across more than ten rounds before the reward's per-error-cell penalty ($-0.05\,k_i$) and sampling-side safeguards (truncating four-error loops and replacing million-token rollouts) drove it back down---dropping runtime errors from 69 to 14 per 100 by round~64, precisely as held-out success accelerated from the mid-30s to \RLSuccesses/100.

\section{Discussion and Limitations}
EmbodyAgent demonstrates that a complete post-training stack---black-box strong-teacher data synthesis, multimodal SFT, and online outcome-based RL---can turn an open-weight 9B VLM that initially solves 0/100 tasks into a reliable closed-loop embodied agent achieving \RLSuccesses/100. Anchoring all three stages on a single primitive-free \emph{think-then-code} interface keeps the VLM inside its native multimodal text-and-code space throughout: Stage~1 synthesizes grounded trajectories from a black-box teacher without needing teacher weights or logits; Stage~2 adapts both vision and language parameters to resolve cold-start contract failure; and Stage~3 optimizes the student VLM directly on its own rollouts, overcoming imitation distribution shift while compressing decision frequency by over $3\times$.

Stage~3 also highlights a structural lesson for post-training executable VLM agents with outcome rewards: because runtime tracebacks re-enter the context, outcome-based RL contains a positive feedback loop on the policy's own runtime errors. Aggregate success rate is a lagging indicator of this pathology---it continued rising for six rounds while runtime errors climbed toward saturation. Per-round failure-category monitoring is therefore essential alongside success tracking; paired with sampling-side safeguards (cell resampling, loop truncation, and memory-outlier replacement) and a small per-error penalty in the reward, optimization recovers cleanly from the error basin without altering reward definitions or test-time evaluation.

Several limitations define the scope for future work. Our evaluation covers one simulation suite (ten LIBERO Spatial tasks) with a 100-episode synthesized corpus and a 9B VLM student; each training stage reflects a single run without multi-seed replication, giving the headline score on 100 test states a binomial standard error of $\pm 3.7$ points (and ten states per task). Task~4 remains challenging at 4/10, showing that outcome-based RL over sparse rewards does not eliminate per-task difficulty ceilings. Moreover, fixing 100 evaluation states enables exact round-by-round comparability at the cost of coarse initial-state coverage. Extending this black-box synthesis $\to$ SFT $\to$ RL post-training pipeline to broader manipulation suites, real-robot execution, and controlled ablations isolating each stage and representation component (explanations, program structure, and observation scheduling) remains natural next work.

\section{Conclusion}
We introduced EmbodyAgent, a full-pipeline post-training framework for embodied VLM agents comprising weight-free black-box teacher data synthesis, multimodal supervised fine-tuning, and group-relative outcome reinforcement learning over one unified think-then-code Python interface. On ten LIBERO Spatial tasks under a strictly separated evaluation protocol, the pipeline elevates an open-weight Qwen3.5-9B VLM from 0/100 (base) to 14/100 (SFT) and \RLSuccesses/100 (RL), while cutting median test model calls from 36 to 11 and removing any ongoing dependence on the black-box teacher. We further diagnosed a self-reinforcing runtime-error loop unique to outcome optimization over executable feedback and provided the monitoring and sampling-side curation protocol that keeps RL stable. These results establish end-to-end post-training over executable programs as a coherent and effective recipe for building self-improving embodied VLM agents.

\bibliographystyle{plain}
\begingroup
\raggedright
\bibliography{references}

@misc{li2025simplevla,
  author = {Haozhan Li and Yuxin Zuo and Jiale Yu and others},
  title = {{SimpleVLA-RL}: Scaling {VLA} Training via Reinforcement Learning},
  year = {2025}, howpublished = {\url{https://arxiv.org/abs/2509.09674}},
  note = {RL infrastructure adapted for this work}
}

@misc{fan2026captain,
  author = {Yijia Fan and Xilin He and Yilei Jiang and Dongming Wu and Jiaxu Wang and Xiangyu Yue},
  title = {{CaPtain}: A Pipeline for Post-Training Code-as-Policy Robot Agents},
  year = {2026}, howpublished = {\url{https://arxiv.org/abs/2610.40799}}
}

@inproceedings{liang2022code,
  author = {Jacky Liang and Wenlong Huang and Fei Xia and Peng Xu and Karol Hausman and Brian Ichter and Pete Florence and Andy Zeng},
  title = {Code as Policies: Language Model Programs for Embodied Control},
  booktitle = {ICRA}, year = {2023}
}

@inproceedings{ahn2022saycan,
  author = {Michael Ahn and Anthony Brohan and Noah Brown and others},
  title = {Do As I Can, Not As I Say: Grounding Language in Robotic Affordances},
  booktitle = {CoRL}, year = {2022}
}

@inproceedings{huang2022inner,
  author = {Wenlong Huang and Fei Xia and Ted Xiao and others},
  title = {Inner Monologue: Embodied Reasoning through Planning with Language Models},
  booktitle = {CoRL}, year = {2022}
}

@inproceedings{singh2023progprompt,
  author = {Ishika Singh and Valts Blukis and Arsalan Mousavian and others},
  title = {{ProgPrompt}: Generating Situated Robot Task Plans Using Large Language Models},
  booktitle = {ICRA}, year = {2023}
}

@inproceedings{huang2023voxposer,
  author = {Wenlong Huang and Chen Wang and Ruohan Zhang and Li Fei-Fei},
  title = {{VoxPoser}: Composable {3D} Value Maps for Robotic Manipulation with Language Models},
  booktitle = {CoRL}, year = {2023}
}

@inproceedings{wang2024codeact,
  author = {Xingyao Wang and Yangyi Chen and Lifan Yuan and others},
  title = {Executable Code Actions Elicit Better Language Model Agents},
  booktitle = {ICML}, year = {2024}
}

@inproceedings{brohan2022rt1,
  author = {Anthony Brohan and Noah Brown and Justice Carbajal and others},
  title = {{RT-1}: Robotics Transformer for Real-World Control at Scale},
  booktitle = {RSS}, year = {2023}
}

@inproceedings{brohan2023rt2,
  author = {Anthony Brohan and Noah Brown and Justice Carbajal and others},
  title = {{RT-2}: Vision-Language-Action Models Transfer Web Knowledge to Robotic Control},
  booktitle = {CoRL}, year = {2023}
}

@inproceedings{octo2024,
  author = {{Octo Team} and others},
  title = {Octo: An Open-Source Generalist Robot Policy},
  booktitle = {RSS}, year = {2024}
}

@misc{black2024pi0,
  author = {Kevin Black and Noah Brown and Danny Driess and others},
  title = {$\pi_0$: A Vision-Language-Action Flow Model for General Robot Control},
  year = {2024}, howpublished = {\url{https://arxiv.org/abs/2410.24164}}
}

@misc{liu2024rdt,
  author = {Songming Liu and Lingxuan Wu and Bangguo Li and others},
  title = {{RDT-1B}: A Diffusion Foundation Model for Bimanual Manipulation},
  year = {2024}, howpublished = {\url{https://arxiv.org/abs/2410.07864}}
}

@inproceedings{kim2024openvla,
  author = {Moo Jin Kim and Karl Pertsch and Siddharth Karamcheti and others},
  title = {OpenVLA: An Open-Source Vision-Language-Action Model},
  booktitle = {CoRL}, year = {2024}
}

@inproceedings{liu2023libero,
  author = {Bo Liu and Yifeng Zhu and Chongkai Gao and Yihao Feng and Qiang Liu and Yuke Zhu and Peter Stone},
  title = {{LIBERO}: Benchmarking Knowledge Transfer for Lifelong Robot Learning},
  booktitle = {NeurIPS}, year = {2023}
}

@inproceedings{wei2022cot,
  author = {Jason Wei and Xuezhi Wang and Dale Schuurmans and others},
  title = {Chain-of-Thought Prompting Elicits Reasoning in Large Language Models},
  booktitle = {NeurIPS}, year = {2022}
}

@misc{zawalski2024ecot,
  author = {Micha{\l} Zawalski and William Chen and Karl Pertsch and others},
  title = {Robotic Control via Embodied Chain-of-Thought Reasoning},
  year = {2024}, howpublished = {\url{https://arxiv.org/abs/2407.08693}}
}

@misc{zhao2025cotvla,
  author = {Qingqing Zhao and Yao Lu and Moo Jin Kim and others},
  title = {{CoT-VLA}: Visual Chain-of-Thought Reasoning for Vision-Language-Action Models},
  year = {2025}, howpublished = {\url{https://arxiv.org/abs/2503.22020}}
}

@misc{shao2024grpo,
  author = {Zhihong Shao and Peiyi Wang and Qihao Zhu and others},
  title = {{DeepSeekMath}: Pushing the Limits of Mathematical Reasoning in Open Language Models},
  year = {2024}, howpublished = {\url{https://arxiv.org/abs/2402.03300}},
  note = {Introduces group-relative policy optimization}
}

@misc{schulman2017ppo,
  author = {John Schulman and Filip Wolski and Prafulla Dhariwal and Alec Radford and Oleg Klimov},
  title = {Proximal Policy Optimization Algorithms},
  year = {2017}, howpublished = {\url{https://arxiv.org/abs/1707.06347}}
}

@misc{deepseek2025r1,
  author = {{DeepSeek-AI} and others},
  title = {{DeepSeek-R1}: Incentivizing Reasoning Capability in {LLMs} via Reinforcement Learning},
  year = {2025}, howpublished = {\url{https://arxiv.org/abs/2501.12948}}
}

@misc{li2026embodiedevo,
  author = {Wenyi Li and Yizhe Chi and Youjie Zheng and others},
  title = {{EmbodiedEvo}: Self-Evolving Coding Agent for Physical Intelligence},
  year = {2026}, howpublished = {\url{https://arxiv.org/abs/2610.18867}}
}

@inproceedings{yao2023react,
  author = {Shunyu Yao and Jeffrey Zhao and Dian Yu and Nan Du and Izhak Shafran and Karthik Narasimhan and Yuan Cao},
  title = {{ReAct}: Synergizing Reasoning and Acting in Language Models},
  booktitle = {ICLR}, year = {2023}
}

@inproceedings{zelikman2022star,
  author = {Eric Zelikman and Yuhuai Wu and Jesse Mu and Noah D. Goodman},
  title = {{STaR}: Bootstrapping Reasoning With Reasoning},
  booktitle = {NeurIPS}, year = {2022}
}

@inproceedings{schick2023toolformer,
  author = {Timo Schick and Jane Dwivedi-Yu and others},
  title = {Toolformer: Language Models Can Teach Themselves to Use Tools},
  booktitle = {NeurIPS}, year = {2023}
}

@inproceedings{shinn2023reflexion,
  author = {Noah Shinn and Federico Cassano and Edward Berman and others},
  title = {Reflexion: Language Agents with Verbal Reinforcement Learning},
  booktitle = {NeurIPS}, year = {2023}
}

@inproceedings{yang2024sweagent,
  author = {John Yang and Carlos Jimenez and Alexander Wettig and others},
  title = {{SWE-agent}: Agent-Computer Interfaces Enable Automated Software Engineering},
  booktitle = {NeurIPS}, year = {2024}
}

@misc{yang2025qwen3,
  author = {An Yang and Anfeng Li and Baosong Yang and others},
  title = {Qwen3 Technical Report},
  year = {2025}, howpublished = {\url{https://arxiv.org/abs/2505.09388}}
}

@inproceedings{jimenez2024swebench,
  author = {Carlos E. Jimenez and John Yang and Alexander Wettig and others},
  title = {{SWE-bench}: Can Language Models Resolve Real-World {GitHub} Issues?},
  booktitle = {ICLR}, year = {2024}
}

@inproceedings{hsieh2023distilling,
  author = {Cheng-Yu Hsieh and Chun-Liang Li and Chih-Kuan Yeh and others},
  title = {Distilling Step-by-Step! Outperforming Larger Language Models with Less Training Data and Smaller Model Sizes},
  booktitle = {Findings of ACL}, year = {2023}
}

@inproceedings{lightman2024verify,
  author = {Hunter Lightman and Szymon Sidor and others},
  title = {Let's Verify Step by Step},
  booktitle = {ICLR}, year = {2024}
}

@inproceedings{le2022coderl,
  author = {Hung Le and Yue Wang and others},
  title = {{CodeRL}: Mastering Code Generation through Pretrained Models and Deep Reinforcement Learning},
  booktitle = {NeurIPS}, year = {2022}
}

@inproceedings{li2023roboflamingo,
  author = {Xinghang Li and Minghuan Liu and Hanbo Zhang and others},
  title = {Vision-Language Foundation Models as Effective Robot Imitators},
  booktitle = {CoRL}, year = {2023}
}

@inproceedings{loshchilov2019adamw,
  author = {Ilya Loshchilov and Frank Hutter},
  title = {Decoupled Weight Decay Regularization},
  booktitle = {ICLR}, year = {2019}
}

@inproceedings{rajbhandari2020zero,
  author = {Samyam Rajbhandari and Jeff Rasley and Olatunji Ruwase and Yuxiong He},
  title = {{ZeRO}: Memory Optimizations toward Training Trillion Parameter Models},
  booktitle = {SC}, year = {2020}
}

@inproceedings{kwon2023vllm,
  author = {Woosuk Kwon and Zhuohan Li and Siyuan Zhuang and others},
  title = {Efficient Memory Management for Large Language Model Serving with {PagedAttention}},
  booktitle = {SOSP}, year = {2023}
}

@inproceedings{yang2024deltanet,
  author = {Songlin Yang and others},
  title = {Parallelizing Linear Transformers with the Delta Rule over Sequence Length},
  booktitle = {NeurIPS}, year = {2024}
}

@misc{yuan2026openrua,
  author = {Jiarui Yuan and Yuxin Zuo and Xiaoyu Huang and others},
  title = {Democratizing Robot-Use Agents: A Unified Harness with Open Models},
  year = {2026}, howpublished = {\url{https://arxiv.org/abs/2611.01560}}
}

@misc{yu2025dapo,
  author = {Qiying Yu and Zheng Zhang and Ruofei Zhu and Yufeng Yuan and Xiaochen Zuo and others},
  title = {{DAPO}: An Open-Source {LLM} Reinforcement Learning System at Scale},
  year = {2025}, howpublished = {\url{https://arxiv.org/abs/2503.14476}}
}

@misc{isola2026robotuse,
  author = {Phillip Isola},
  title = {Robot-Use Agents},
  year = {2026}, howpublished = {\url{https://web.mit.edu/phillipi/www/writing/robot-use-agents.html}}
}

@misc{chu2026openrua,
  author = {Zhaoyang Chu and Earl T. Barr and Claire Le Goues and others},
  title = {{OpenRUA}: Robot-Use Agents Are Zero-Shot Visuomotor Policies},
  year = {2026}, howpublished = {\url{https://arxiv.org/abs/2610.02459}}
}

@misc{zhang2026unexpected,
  author = {Wenbo Zhang and Kaixuan Wang and Yutao Ouyang and others},
  title = {An Unexpected Robot Policy: Early Evaluations of {GPT-6 Astra} on {RoboDojo} and Beyond},
  year = {2026}, howpublished = {\url{https://arxiv.org/abs/2609.24170}}
}

@misc{qu2026roboharness,
  author = {Bince Qu and Wei Chen and Bo Zhang},
  title = {{RoboHarness}: A Simple Harness Outperforms {VLA} and World Action Models},
  year = {2026}, howpublished = {\url{https://arxiv.org/abs/2610.14018}}
}

@misc{li2026k1,
  author = {Zexi Li and Yehang Zhang and Wenqian Li and others},
  title = {{Robo-Harness K1}: Harnessing Robot-Use Agents via Perception Augmentation},
  year = {2026}, howpublished = {\url{https://arxiv.org/abs/2609.29389}}
}

@misc{chen2026showharness,
  author = {Yanzhe Chen and Zechen Bai and Zhijun Cao and others},
  title = {{Show-Harness}: Just a {VLM} Agent Can Play Robots},
  year = {2026}, howpublished = {\url{https://arxiv.org/abs/2609.10522}}
}

@misc{guo2026robodawn,
  author = {Meng-Hao Guo and Zhe-Han Mo and Jia-Jun Wang and others},
  title = {Transferring the Intelligence of {VLMs} to Robotic Control},
  year = {2026}, howpublished = {\url{https://arxiv.org/abs/2609.22966}}
}

@misc{si2026fewertokens,
  author = {Ruiyang Si and Jianxin Bi and Shunyu Yang and others},
  title = {Fewer Tokens, Better Action: {GPT-6 Astra} Robot Agents with 14\% Higher Success Rate but 65\% Fewer Tokens},
  year = {2026}, howpublished = {\url{https://arxiv.org/abs/2610.01939}}
}

@misc{agashe2025agents2,
  author = {Saaket Agashe and Kyle Wong and Vincent Tu and others},
  title = {Agent {S2}: A Compositional Generalist-Specialist Framework for Computer Use Agents},
  year = {2025}, howpublished = {\url{https://arxiv.org/abs/2504.00906}}
}

@misc{wang2025uitars2,
  author = {Haoming Wang and Haoyang Zou and Huatong Song and others},
  title = {{UI-TARS-2} Technical Report: Advancing {GUI} Agent with Multi-Turn Reinforcement Learning},
  year = {2025}, howpublished = {\url{https://arxiv.org/abs/2509.02544}}
}

@inproceedings{driess2023palme,
  author = {Danny Driess and Fei Xia and Mehdi S. M. Sajjadi and others},
  title = {{PaLM-E}: An Embodied Multimodal Language Model},
  booktitle = {ICML}, year = {2023}
}

@misc{zhang2026harnessvla,
  author = {Yixian Zhang and Huanming Zhang and Feng Gao and others},
  title = {Harness {VLA}: Steering Frozen {VLAs} into Reliable Manipulation Primitives via Memory-Guided Agents},
  year = {2026}, howpublished = {\url{https://arxiv.org/abs/2607.08448}}
}
\endgroup
\appendix
\section{Reproducibility Checklist}
\label{app:protocol}
\begin{center}
\scriptsize
\begin{tabular}{@{}>{\raggedright\arraybackslash}p{0.30\linewidth}>{\raggedright\arraybackslash}p{0.64\linewidth}@{}}
\toprule
Item & Protocol \\
\midrule
Suite & LIBERO Spatial, tasks 0--9 \cite{liu2023libero} \\
Environments & Created from official LIBERO BDDL task definitions \\
Initial states & Official pre-generated state pool, pinned by a manifest (\texttt{official\_state\_split\_v1}) with per-state SHA-256 hashes \\
State split & Per task: 30 train / 10 validation / 10 test states; training and validation feed the teacher and RL sampling; the 100 test states are scored after every round for monitoring, and the reported checkpoint is the best score observed under this monitoring \\
Policy tool & Single \texttt{execute\_python} tool; no primitives beyond \texttt{env.observation()} and \texttt{env.action($a\in[-1,1]^7$)} \\
Observation & Primary + wrist images and proprioception, supplied only when the program requests them; no automatic post-code observation; no privileged state \\
Success label & Environment BDDL success predicate; failure category recorded per episode \\
Episode budget & 2{,}000 environment steps, 150 model calls, 600-second code-block timeout, or voluntary \texttt{STOP\_EPISODE}; identical for training rollouts and evaluation \\
Teacher corpus & Astra Medium (frontier VLM, medium reasoning effort) through the same tool and contract; 100 successful episodes (ten per task), all on train-split states; 1{,}475 assistant-response targets, each pairing an explanation with a program \\
Student SFT & Qwen3.5-9B\cite{yang2025qwen3}; vision encoder and language parameters both trained; assistant-token loss on explanation, code, and format tokens; ten epochs; RL starts after the distilled checkpoint passes output-format and tool-call validation on held-out states \\
RL objective & Shaped success reward (\texttt{binary\_success\_minus\_cell\_errors\_v1}): binary outcome minus a capped $0.05$ per-error-cell penalty, fast-fail floor $-0.2$; group size $G=8$; 32 valid groups per round (uniform-binary-outcome groups discarded, at most 160 attempted); episode-level group-relative advantage; infrastructure failures resampled up to three times and excluded from reward statistics \\
RL optimization & Temperature-one sampling; AdamW on all parameters including the vision encoder; ZeRO-3 sharding; constant learning rate $2\times10^{-6}$; no KL term; asymmetric clip $(\varepsilon_{\mathrm{lo}},\varepsilon_{\mathrm{hi}})=(0.2,0.28)$ \\
Compute & RL training on 16 PPU accelerators (ZeRO-3); policy inference for training rollouts and evaluation served by vLLM; training accelerated by flash-linear-attention kernels for the gated-delta-rule layers; SFT gate: RL starts only after the distilled checkpoint passes output-format and tool-call validation on held-out states \\
RL schedule & 70 rounds configured; the reported run is in progress within this schedule; every round's checkpoint and per-round test scores are preserved \\
Training safeguards & Sampling-only: (i) unparseable/non-compiling cells resampled up to twice; (ii) episode truncated after four runtime-error cells; (iii) failed rollouts exceeding $10^6$ context tokens replaced up to three times per group with advantages recomputed. None applies at evaluation \\
Evaluation & All 100 test states after every RL round; greedy decoding; full budgets; no resampling, truncation, or replacement; per-episode failure categories logged \\
\bottomrule
\end{tabular}
\end{center}

\subsection{Execution and training details}
The runtime maintains a persistent Python namespace across each episode. Tool feedback contains captured stdout, stderr, exceptions, the last expression's value, and any requested images. An exception does not undo actions already executed by earlier statements of the same cell. The first model turn receives no automatically supplied image, and the harness does not append an image after a program unless the program requested one. The interface exposes robot proprioception but no object poses, contact oracles, or simulator internals. The Python process is not a security sandbox.

At update time, only newly generated assistant tokens contribute to the policy loss; environment feedback and preceding history remain masked. The episode-level advantage is shared across all explanation and program tokens of the episode. Rollouts store the exact sampled tokens and their old-policy log-probabilities, and training consumes exactly those tokens.

Two numerical guards are worth recording for replication under ZeRO-3. First, an initialization audit found that the visual rotary-frequency buffer is materialized in reduced precision under sharded initialization; rebuilding this non-persistent buffer in FP32 restores the standalone inference construction without changing any trained weights. Second, the frozen vision encoder's patch convolution is expressed as its algebraically equivalent linear projection during ZeRO parameter gathering, which removes a standalone-versus-distributed numerical divergence observed at the update-consistency guard; sampled likelihoods, rewards, masks, and guard thresholds are unchanged by this reformulation.

\end{document}